\documentclass{article}

\usepackage[preprint]{neurips_2026}

\usepackage[utf8]{inputenc} 
\usepackage[T1]{fontenc}    
\usepackage[hyperfootnotes=false]{hyperref} 
\usepackage{url}            
\usepackage{booktabs}       
\usepackage{amsfonts}       
\usepackage{nicefrac}       
\usepackage{microtype}      
\usepackage{xcolor}         

\usepackage{times}
\usepackage{epsfig}
\usepackage{graphicx}
\usepackage{amsmath}
\usepackage{amssymb}
\usepackage{xcolor}         
\usepackage{float}
\usepackage{caption}
\usepackage{amsfonts,amssymb}
\usepackage{mathrsfs}
\usepackage{tabularx}
\usepackage{wrapfig}
\usepackage{verbatim}
\usepackage{dsfont}
\usepackage{tablefootnote}
\usepackage[stable]{footmisc}
\usepackage{booktabs}
\usepackage{multicol}
\usepackage{multirow}
\usepackage{color}
\usepackage{subcaption}
\usepackage{pythonhighlight}
\usepackage{algpseudocode}
\usepackage{algorithm}
\usepackage{physics}

\renewcommand{\paragraph}[1]{\smallskip\noindent\textbf{#1}}

\definecolor{sim}{RGB}{128,0,128}
\definecolor{interp}{RGB}{255, 169, 0}
\definecolor{mytbcol}{RGB}{175,227,246}

\usepackage{xcolor}
\usepackage{color, colortbl}
\usepackage{xcolor}
\usepackage{adjustbox}

\renewcommand{\paragraph}[1]{\smallskip\noindent\textbf{#1}}

\definecolor{tabfirst}{rgb}{1, 0.7, 0.7} 
\definecolor{tabsecond}{rgb}{1, 0.85, 0.7} 
\definecolor{tabthird}{rgb}{1, 1, 0.7} 

\usepackage{hyperref}
\hypersetup{
    colorlinks=true,
    linkcolor=black,
    citecolor=green,
    urlcolor=blue
}

\usepackage{xspace}

\newcommand{\method}{\textit{FlashRelight}\xspace}

\usepackage{wrapfig}
\usepackage{booktabs}
\usepackage{graphicx}
\usepackage{pifont}
\newcommand{\cmark}{\ding{51}}

\usepackage[most]{tcolorbox}
\usepackage{xcolor}

\tcbset{
  promptbox/.style={
    enhanced,
    breakable,
    colback=blue!4,
    colframe=blue!60!black,
    coltitle=white,
    colbacktitle=blue!60!black,
    fonttitle=\bfseries,
    arc=3mm,
    boxrule=0.8pt,
    left=1mm,
    right=1mm,
    top=1mm,
    bottom=1mm,
    listing only,
    listing options={
      basicstyle=\ttfamily\tiny,
      breaklines=true,
      columns=fullflexible,
      keepspaces=true
    }
  }
}

\title{FlashRelight: Portrait Video Relighting \\ with Dynamic Lighting}
\author{
  \textbf{Jiahui Sheng$^1$\thanks{Work done during an internship at vivo.\quad\textsuperscript{\(\dagger\)}Project leader.\quad\textsuperscript{\(\ddagger\)}Corresponding authors.} \quad
  Le Li$^{2\dagger}$ \quad
  Limin Lin$^2$ \quad
  Jiahao Li$^2$ \quad
  Wansen Feng$^2$} \\
  \textbf{Hong Gu$^2$ \quad
  Shuhan Chen$^1$ \quad
  Xiaorun Li$^{1\ddagger}$ \quad
  Rui Hu$^{2\ddagger}$} \\
  $^1$ Zhejiang University, China \\
  $^2$ vivo BlueImage Lab, vivo Mobile Communication Co., Ltd., China
}

\newif\ifincludeadditionaldomains
\includeadditionaldomainstrue

\begin{document}

\maketitle
\begin{figure}[h]
    \centering
    \includegraphics[width=0.76\linewidth]{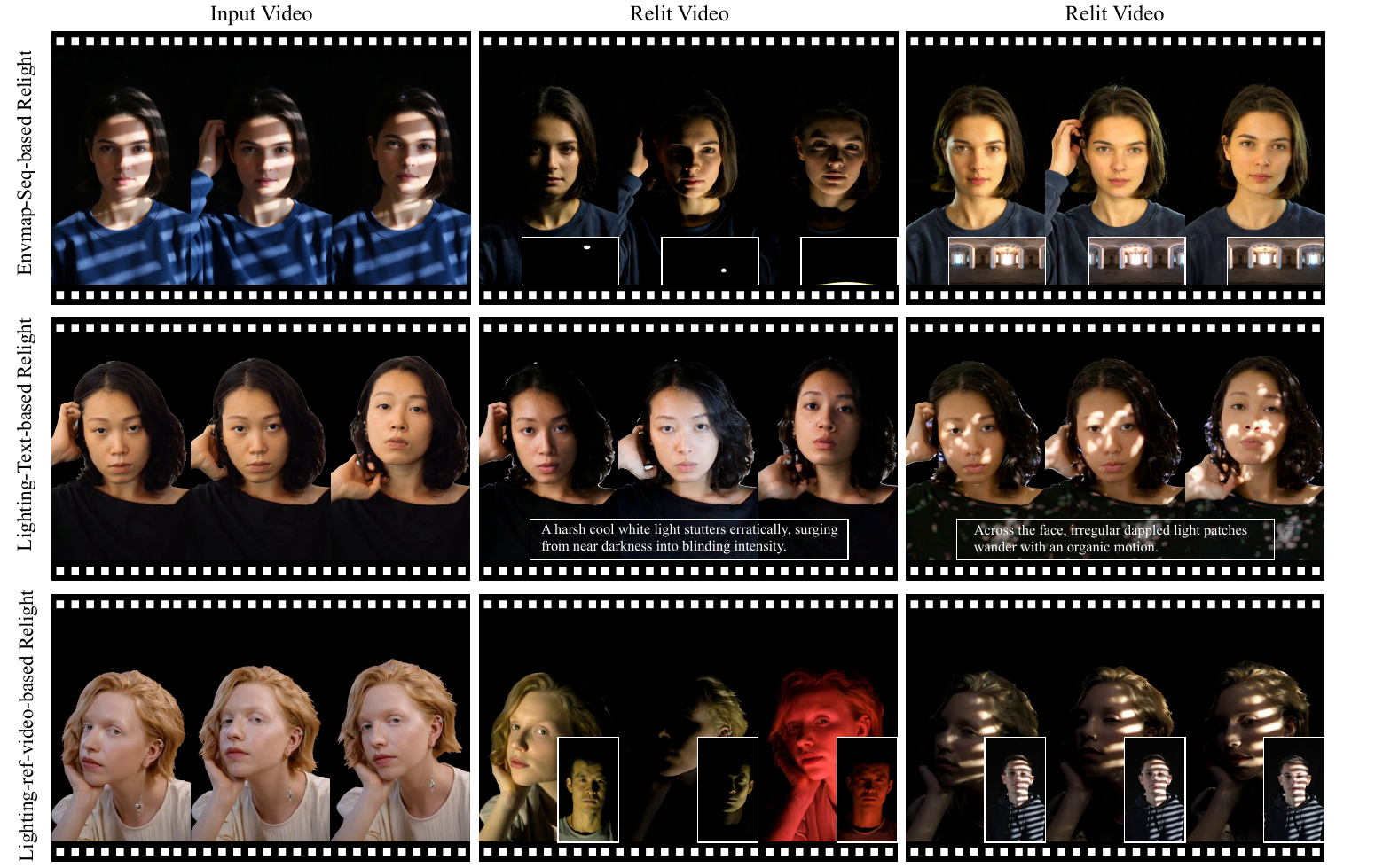}
    \caption{
     FlashRelight enables portrait video relighting with dynamic lighting through three lighting control modalities: envmap sequences, lighting text, and lighting reference videos. 
    }
    \label{fig:Teaser}
\end{figure}

\noindent\begin{minipage}{\linewidth}
\begin{abstract}

Portrait video relighting aims to modify the illumination of a portrait video while preserving photorealism and temporal stability.
The task becomes particularly challenging under dynamic lighting, where illumination direction, intensity, and color vary over time.
To tackle this problem, we introduce \method, a novel two-stage framework for dynamic portrait video relighting. 
\method\ first suppresses the source illumination through delighting, and then synthesizes the target dynamic lighting on an illumination-neutral portrait video. 
\method\ supports three complementary control modalities: envmap sequences for explicit frame-level lighting control, dynamic lighting text for semantic illumination editing, and lighting reference videos for visual lighting transfer. 
Training follows a low-cost hybrid strategy that combines static one-light-at-a-time (OLAT) data with generated dynamic-lighting portrait videos, covering both controlled illumination variation and realistic portrait motion. 
Extensive experiments show that \method\ achieves temporally coherent and photorealistic dynamic relighting while enabling flexible cross-video lighting transfer.
Code is available at \url{https://github.com/vivoCameraResearch/FlashRelight}.
\end{abstract}

\end{minipage}
\par

\section{Introduction}
\label{SecIntro}


\begin{figure*}[t]
    \centering
    \includegraphics[width=\linewidth]{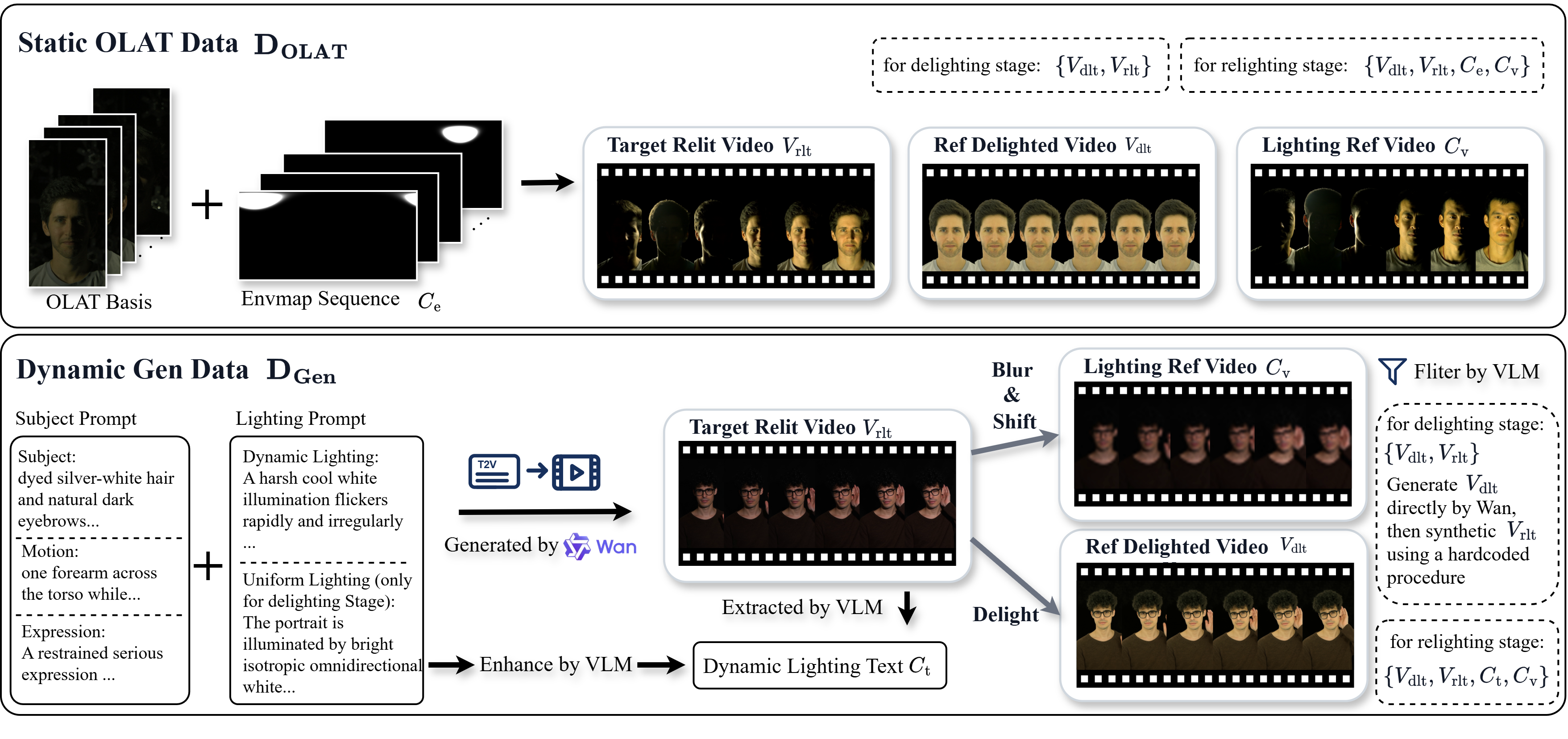}
    \caption{\textbf{Data curation pipeline.} Hybrid Dynamic Portrait Video Dataset $\mathbf{D}_{\mathrm{Hybrid}}$ includes two data sources: \emph{Static OLAT Data} $\mathbf{D}_{\mathrm{OLAT}}$ and \emph{Dynamic Gen Data} $\mathbf{D}_{\mathrm{Gen}}$. }
    \label{fig:data_pipeline}
\end{figure*}

Dynamic lighting is one of the most expressive tools in visual storytelling, shaping the mood, style, and visual impact of portrait videos. 
Effects such as strobing flashes, drifting shadows, moving highlights, and rapid illumination transitions are widely used in professional production, yet remain difficult to create without specialized equipment or extensive post-processing. 
Portrait video relighting offers a computational alternative by modifying the illumination of an existing portrait video while preserving identity, motion, and visual realism.

Existing methods have made notable progress in portrait video relighting, ranging from real-time approaches~\cite{NVPR,relight-cai,ReliTalk} that process portrait videos frame by frame to video diffusion model-based methods~\cite{RelightVid,LuxPostFacto} that improve temporal coherence and visual fidelity. 
However, most prior works focus on static illumination or only mild lighting variations~\cite{NVPR,LuxPostFacto,SwitchLight}. 
In real-world videos and practical editing scenarios, lighting may change over time in direction, intensity, color, and spatial distribution, producing complex dynamic effects such as alternating colored lights, localized flicker, moving highlights, and evolving shadows. 
Dynamic portrait video relighting under complex time-varying illumination therefore remains an important but insufficiently explored problem.

Training such a model requires paired data under dynamic lighting conditions, which is difficult to obtain at scale. 
Conventional data pipelines rely on dynamic OLAT capture~\cite{NVPR} or fully rendered paired data~\cite{RelightMaster,SynthLight}; however, dynamic OLAT requires expensive light-stage setups and scales poorly to unconstrained portrait motion, while rendered data is costly to construct and often suffers from a non-negligible synthetic-to-real domain gap. 
Static OLAT data is more affordable and provides rich relighting supervision, but it is limited to static portraits and cannot capture the interaction between portrait dynamics and time-varying illumination~\cite{POLAR,3DPR}. 
Leveraging recent advances in large-scale video generation models~\cite{wan2025,CogVideoX,HunyuanVideo}, we adopt a low-cost hybrid data strategy that combines static OLAT data with generated dynamic-lighting portrait videos, providing scalable supervision for both controlled illumination variation and realistic portrait motion.

Building on this hybrid data strategy, we propose \method, a unified framework for portrait video relighting under dynamic lighting conditions. 
Conceptually, \method\ treats dynamic portrait relighting not as frame-wise illumination editing, but as controllable video generation under time-varying lighting conditions. 
To reduce the ambiguity caused by unknown source illumination, \method\ adopts a practical two-stage formulation: a delighting model first suppresses the source lighting to obtain an approximately illumination-neutral portrait video, and a DiT-based relighting model then synthesizes a relit video under the target dynamic illumination by leveraging a pre-trained video generative prior~\cite{ho2022videodiffusionmodels,wan2025}. 
This design allows the model to preserve portrait identity and motion while generating temporally coherent lighting effects over time.

\method\ supports three complementary lighting controls: envmap sequences, dynamic lighting text, and lighting reference videos.
Envmap sequences provide explicit frame-level lighting control~\cite{UniRelight,DiffusionRenderer}, dynamic lighting text provides a semantic conditioning signal~\cite{Text2Relight}, and lighting reference videos provide direct visual guidance for cross-video lighting transfer.
These modalities offer different ways to specify target lighting, with envmap sequences and lighting reference videos serving as our primary means of dynamic lighting control.

In summary, our contributions are as follows:
\begin{itemize}
    \item We propose a novel method, \method,  for portrait video relighting under dynamic lighting conditions, which produces high-fidelity relit videos while maintaining strong temporal stability.
    
    \item \method\ supports multiple lighting control modalities, including envmap sequences, dynamic lighting text, and lighting reference videos.
    Reference-video conditioning further enables direct lighting transfer across portrait videos.
    
    \item We propose a scalable hybrid supervision strategy that combines static OLAT data with generated dynamic-lighting portrait videos, enabling scalable training without large-scale dynamic light-stage capture or fully rendered paired datasets.
\end{itemize}

\section{Related Work}
\label{SecRelated}

\paragraph{Portrait Video Relighting.}
Portrait relighting has been studied for decades, with light-stage systems capturing OLAT reflectance fields and recombining them to synthesize novel illumination~\cite{ReflectanceFieldFace}.
Follow-up works extend this paradigm to dynamic performances using time-multiplexed illumination~\cite{PerformanceRelighting,FlowedReflectanceFields}, reflectance transfer~\cite{ReflectanceTransfer}, and volumetric capture~\cite{TheRelightables,DeepRelightableTextures}.
Although these methods can produce high-quality relit performances, they require specialized capture systems and per-subject acquisition, limiting their applicability to unconstrained portrait videos.

Learning-based methods reduce this capture burden by training relighting models from multi-illumination data.
Representative image-based approaches achieve realistic portrait relighting with learned light transport~\cite{SingleImagePortraitRelighting,TotalRelighting}, explicit 3D priors~\cite{SwitchLight,HoloRelighting,Lite2Relight}, or diffusion models~\cite{CLD,DiFaReLi,SynthLight}.
However, these methods are primarily designed for still images and often suffer from flickering when applied frame by frame to videos.
Video-specific methods improve temporal consistency through consistency modeling~\cite{NVPR}, 3D-aware representations~\cite{relight-cai}, or relightable avatar reconstruction~\cite{DeepRelightableAppearance,URAvatar}.
Nevertheless, they often rely on simplified lighting assumptions, per-subject reconstruction, or scarce dynamic OLAT data.

Recent work explores conditional video diffusion for portrait performance relighting.
Lux Post Facto trains a video diffusion model for portrait video relighting under a static environment map, showing strong photorealism and temporal stability~\cite{LuxPostFacto}.
Different from prior portrait relighting methods that mainly focus on static or mildly varying illumination, \method\ targets dynamic portrait relighting under more complex time-varying illumination, which further supports cross-video lighting transfer.

\paragraph{Video Diffusion Models.}
Video diffusion models have recently shown strong capability in generating realistic and temporally coherent videos from text or image conditions~\cite{MakeAVideo,AlignYourLatents,SVD,wan2025}.
Beyond text-to-video synthesis, recent works improve controllability by conditioning generation on signals such as poses~\cite{MagicPose,MagicAnimate}, image prompts~\cite{IPAdapter}, depth maps, and other structural cues~\cite{ControlNet,UniControlNet}.
These advances make video diffusion models powerful priors for video editing tasks that require both photorealism and temporal consistency.

Recent relighting methods also exploit video diffusion priors for temporally coherent lighting editing.
DiffusionRenderer studies neural inverse and forward rendering with video diffusion models~\cite{DiffusionRenderer}, while Light-A-Video, RelightVid, LightCtrl, and UniLumos study video or unified image-video relighting with diffusion-based models~\cite{LightAVideo,RelightVid,LightCtrl,UniLumos}.
Similarly, \method\ leverages a DiT-based video generative prior for dynamic portrait video relighting, where the target illumination evolves over time and can be specified through multiple lighting controls.

\paragraph{Lighting Control and Transfer.}
The choice of lighting representation largely determines the controllability and usability of relighting systems.
Explicit representations, such as environment maps or HDR lighting, provide physically meaningful control and have been widely used in portrait and video relighting~\cite{TotalRelighting,SwitchLight,SynthLight,LuxPostFacto,DiffusionRenderer,UniRelight}.
However, they are difficult for non-expert users to edit directly.
Recent diffusion-based methods improve usability with text prompts, background guidance, or structured lighting attributes~\cite{ICLight,Text2Relight,UniLumos,TokenLight}, while reference-driven methods specify the desired illumination through another image or video~\cite{TransLight,LightAVideo,RelightVid}.
These interfaces are intuitive, but most existing methods still focus on a single control form and are primarily designed for static lighting.

In contrast, portrait video relighting with dynamic lighting requires the target illumination to be time-varying.
\method\ unifies complementary dynamic lighting controls in one framework: environment-map sequences provide precise frame-level control, dynamic lighting text offers a semantic interface, and lighting reference videos enable cross-video lighting transfer. 
This unified design supports both explicit dynamic relighting and flexible cross-video lighting transfer within a single portrait video relighting framework.


\begin{figure*}[t]
    \centering
    \includegraphics[width=\linewidth]{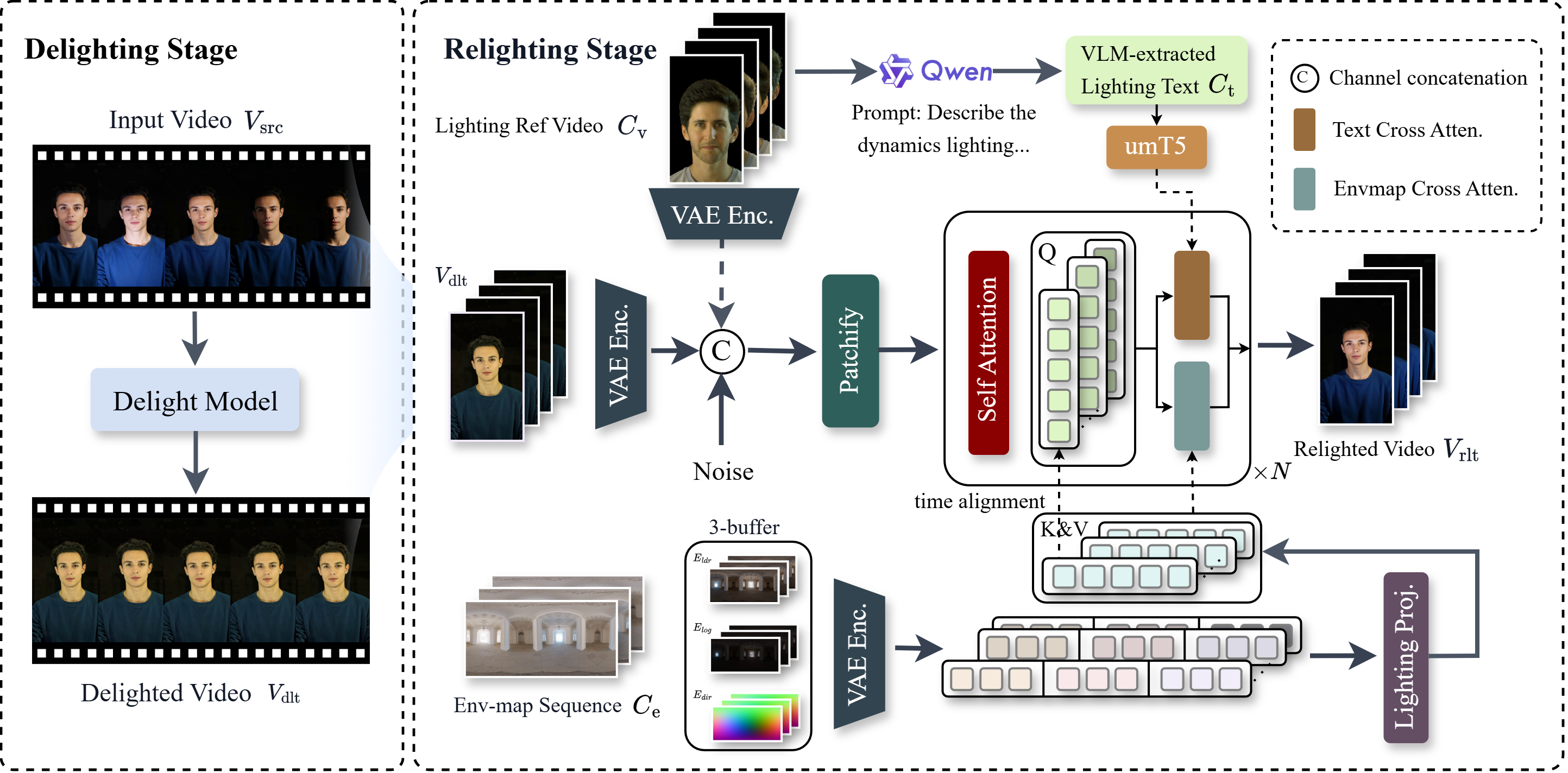}
    \caption{\textbf{Overview of \method.}
    Our framework consists of a delighting stage and a relighting stage.
    The delighting model suppresses source illumination and shadows from the input video, producing an illumination-neutral portrait video.
    The delighted video, together with one active dynamic lighting condition, is then used by the relighting model to synthesize the final relit video.}
    \label{fig:overview}
\end{figure*}

\section{Method}

\subsection{Overview}

\method\ formulates dynamic portrait video relighting as a conditional video generation problem with a delighting-relighting decomposition~\cite{TotalRelighting,HoloRelighting,LightPainter,LuxPostFacto}. 
Given a source portrait video $V_{\mathrm{src}}$, our goal is to generate a relit video that preserves the subject identity, appearance, expression, and motion, while following a target dynamic lighting condition $c$.

We denote by $V_{\mathrm{app}}$ the ideal illumination-neutral appearance video of the source portrait. 
Since $V_{\mathrm{app}}$ is not directly observable, we use a delighting model $\mathcal{D}$ to estimate it:
\begin{equation}
    V_{\mathrm{dlt}} = \mathcal{D}(V_{\mathrm{src}}) \approx V_{\mathrm{app}},
\end{equation}
where $V_{\mathrm{dlt}} \in \mathbb{R}^{T \times C \times H \times W}$ serves as a practical appearance carrier with the source illumination suppressed.

Conditioned on this estimated appearance, relighting is modeled as video generation under the target lighting control:
\begin{equation}
    V_{\mathrm{rlt}} \sim p_{\theta}(V \mid V_{\mathrm{dlt}}, c),
\end{equation}
where the lighting condition $c \in \{C_{\mathrm{e}}, C_{\mathrm{t}}, C_{\mathrm{v}}\}$ can be instantiated as an environment-map sequence, dynamic lighting text, or a lighting reference video. 
This formulation decouples appearance estimation from lighting synthesis, allowing the relighting model to focus on generating the desired illumination dynamics. 
Built upon a pre-trained DiT-based video generative model~\cite{wan2025}, \method\ further benefits from its spatiotemporal prior for temporally coherent video generation.

Next, we introduce the two-stage relighting model, followed by our OLAT-Gen hybrid training data.

\subsection{Two-stage Relighting Model}

\paragraph{Delighting Stage.}
The source video $V_{\mathrm{src}} \in \mathbb{R}^{T \times C \times H \times W}$ contains portrait appearance entangled with unknown source illumination. 
Direct one-stage relighting must implicitly remove the original lighting while synthesizing the target lighting, which is ambiguous under complex or temporally varying illumination. 
Therefore, the delighting stage estimates $V_{\mathrm{dlt}}$ as an approximation of the ideal appearance video $V_{\mathrm{app}}$. 
We do not require $V_{\mathrm{dlt}}$ to be a physically exact intrinsic decomposition; it only needs to provide a stable appearance representation that preserves identity and motion for subsequent relighting. The source-video latent is concatenated with the noisy target latent along the channel dimension to condition the delighting model.

\paragraph{Relighting Stage.}
As shown in Fig.~\ref{fig:overview}, the relighting model $\mathcal{R}$ is built upon a pre-trained DiT-based video generative model~\cite{wan2025}. 
Given the delighted video $V_{\mathrm{dlt}}$ and a target lighting condition $c$, it generates the relit portrait video:
\begin{equation}
    V_{\mathrm{rlt}} = \mathcal{R}(V_{\mathrm{dlt}}, c).
\end{equation}
The latent of the delighted video is concatenated with the noisy target latent along the channel dimension at each denoising step, providing temporally aligned appearance and motion guidance.
The lighting condition $c$ provides the target illumination, supporting three control signals. 
Unlike frame-wise relighting, the model generates the output jointly across space and time, inheriting temporal consistency from the underlying video generative prior.

\paragraph{Unified Dynamic Lighting Controls.}
As shown in Fig.~\ref{fig:overview}, \method\ supports three dynamic lighting controls: envmap sequences, lighting text, and lighting reference videos, offering different trade-offs between precise frame-level control and flexible semantic or reference-based editing.

\noindent\textit{Environment-map sequence.}
Environment-map sequences $C_{\mathrm{e}}$ provide the most explicit frame-level lighting control in our framework. 
Following prior works~\cite{DiffusionRenderer,UniRelight}, we encode each HDR environment map $E_t$ using three complementary buffers: an LDR panorama $E_t^{\mathrm{ldr}} \in \mathbb{R}^{3 \times H' \times W'}$, a log-intensity map $E_t^{\mathrm{log}} \in \mathbb{R}^{3 \times H' \times W'}$, and a directional encoding $E_t^{\mathrm{dir}} \in \mathbb{R}^{3 \times H' \times W'}$. These buffer sequences are passed through the VAE encoder to obtain compact spatiotemporal lighting latents, which are then concatenated and projected into envmap tokens $Z$.

To preserve frame-level correspondence, we perform token-level temporal alignment before envmap cross-attention. For video tokens at temporal index $t$ in the latent sequence, attention is computed only with envmap tokens from the same temporal index:
\begin{equation}
    \mathrm{Attn}_{\mathrm{env}}(X_t, Z_t)
    =
    \mathrm{Softmax}
    \left(
    \frac{Q_t K_t^\top}{\sqrt{d}}
    \right)V_t,
\end{equation}
where $Q_t$ is projected from the self-attention features, and $K_t,V_t$ are projected from the aligned envmap tokens $Z_t$. 

\noindent\textit{Dynamic lighting text.}
Dynamic lighting text $C_{\mathrm{t}}$ provides semantic guidance for dynamic lighting conditions, such as moving lights, color transitions, intensity changes, and flicker.
The lighting text is encoded by T5 and injected through text cross-attention.
At inference, this branch accepts either a user-provided lighting prompt or a VLM-extracted lighting description from a reference video as the text condition.

\noindent\textit{Lighting reference video.}
A lighting reference video $C_{\mathrm{v}}$ provides visual evidence of the desired illumination dynamics for cross-video lighting transfer. We encode it as a dense spatiotemporal condition and fuse it with the relighting input through channel-wise concatenation at the latent input layer. This enables \method\ to perform more fine-grained and direct dynamic lighting transfer.


\subsection{Hybrid Training Data}
\label{sec:hybrid_training_data}

As shown in Fig.~\ref{fig:data_pipeline}, our Hybrid Dynamic Portrait Video Dataset $\mathbf{D}_{\mathrm{Hybrid}}$ combines two complementary data sources: \emph{Static OLAT Data} $\mathbf{D}_{\mathrm{OLAT}}$ and \emph{Dynamic Gen Data} $\mathbf{D}_{\mathrm{Gen}}$. 
Static OLAT data provides controlled and fine-grained lighting supervision, while Dynamic Gen Data scales the distribution to moving subjects and temporally varying illumination.

\paragraph{Static OLAT Data $\mathbf{D}_{\mathrm{OLAT}}$.}
Static OLAT Data is synthesized by linearly combining OLAT basis images with envmap sequences. 
We collect OLAT basis images from FaceOLAT, an open-source portrait OLAT dataset, using 60 subjects, each with 2--4 expressions.
Our lighting set includes 600 synthetic environment-map sequences with diverse lighting dynamics and 200 real environment maps collected from Poly Haven~\cite{PolyHaven}. 
The real environment maps are further augmented by rotation and temporal transformations to construct dynamic environment-map sequences.
This results in 24,000 paired videos in total. Each data pair is denoted as $\{V_{\mathrm{dlt}}, V_{\mathrm{rlt}}, C_{\mathrm{e}}, C_{\mathrm{v}}\}$, where $V_{\mathrm{rlt}}$ is the synthetic video under envmap sequence, $V_{\mathrm{dlt}}$ is synthesized under a uniform illumination map and serves as the illumination-neutral target, $C_{\mathrm{e}}$ is the envmap sequence, and $C_{\mathrm{v}}$ is the lighting reference video, which is the synthetic video under the same envmap sequence with another subject.

\paragraph{Dynamic Gen Data $\mathbf{D}_{\mathrm{Gen}}$.}
Dynamic Gen Data is synthesized using Wan2.2~\cite{wan2025} to complement Static OLAT Data with moving subjects and temporally varying illumination.
Because the delighting and relighting stages require different supervision, we construct two stage-specific variants, $\mathbf{D}_{\mathrm{Gen}}^{\mathrm{dlt}}$ and $\mathbf{D}_{\mathrm{Gen}}^{\mathrm{rlt}}$, which are detailed below.

Each relighting pair in $\mathbf{D}_{\mathrm{Gen}}^{\mathrm{rlt}}$ is denoted as 
$\{V_{\mathrm{dlt}}, V_{\mathrm{rlt}}, C_{\mathrm{t}}, C_{\mathrm{v}}\}$, 
where $V_{\mathrm{rlt}}$ is the dynamic-lighting video, $V_{\mathrm{dlt}}$ is its delighted counterpart, $C_{\mathrm{t}}$ is the dynamic lighting text, and $C_{\mathrm{v}}$ is the lighting reference video. 
The text condition $C_{\mathrm{t}}$ is built from both \emph{VLM-enhanced lighting text} and \emph{VLM-extracted lighting descriptions} using Qwen3.5-27B~\cite{qwen3.5}. 
To construct the lighting reference condition while reducing content leakage, we obtain $C_{\mathrm{v}}$ from $V_{\mathrm{rlt}}$ using strong Gaussian blurring, spatial shifting, and intensity scaling.
These operations suppress identity and structural details while preserving low-frequency illumination dynamics. 
To ensure data quality, we use a VLM-based filter to remove samples with impure backgrounds, incorrect lighting dynamics, or visible light sources. 
In total, Dynamic Gen Data contains over 10,000 paired videos. 
Detailed VLM prompts are provided in Appendix~\ref{VLM_prompt}.

\begin{figure*}[t]
    \centering
    \includegraphics[width=\linewidth]{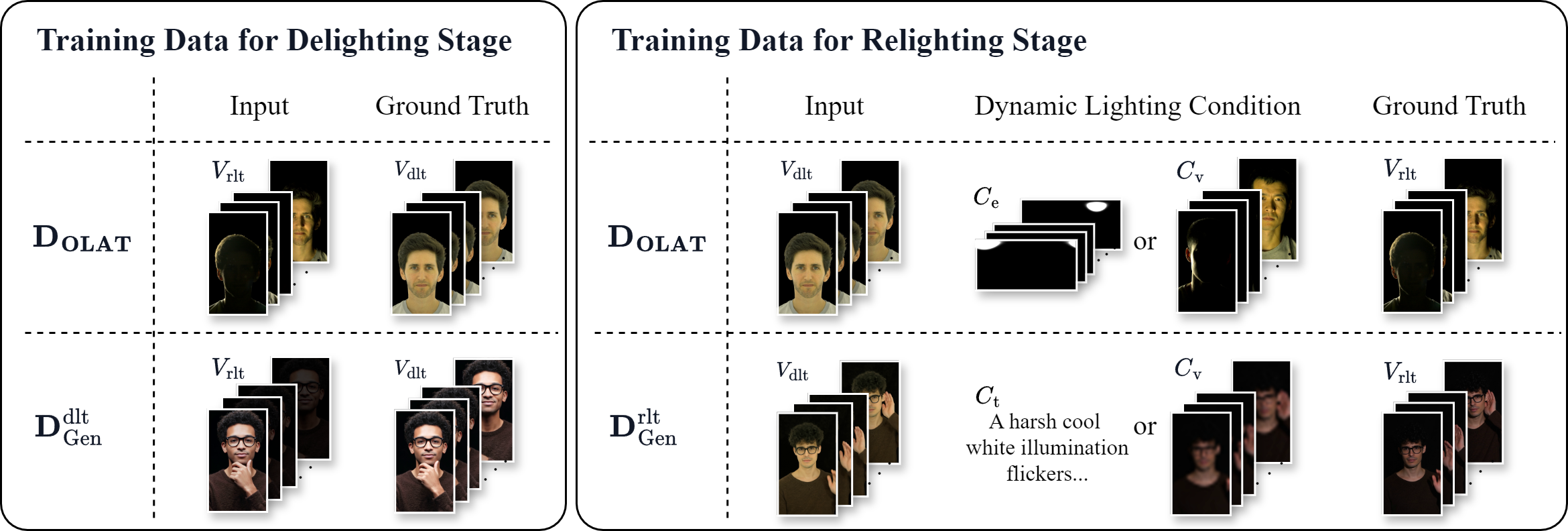}
    \caption{\textbf{Training data for the two stages.} Both the delighting stage and the relighting stage are trained with hybrid data. However, the Dynamic Gen Data $\mathbf{D}_{\mathrm{Gen}}$ used for training the delighting model and that used for training the relighting model are constructed in different ways. }
    \label{fig:two_stage_training_data}
\end{figure*}

\paragraph{Training Data for Delighting Stage.}
As shown in Fig.~\ref{fig:two_stage_training_data}, when training the delighting model, we use paired videos
$(V_{\mathrm{rlt}}, V_{\mathrm{dlt}})$ from both $\mathbf{D}_{\mathrm{OLAT}}$ and $\mathbf{D}_{\mathrm{Gen}}$,
where $V_{\mathrm{rlt}}$ is used as the input and $V_{\mathrm{dlt}}$ serves as the ground truth.
It is worth noting that, for the delighting stage, $V_{\mathrm{dlt}}$ in $\mathbf{D}_{\mathrm{Gen}}$
is directly generated as an illumination-neutral video, while $V_{\mathrm{rlt}}$ is obtained by applying hard-coded dynamic lighting transformations to $V_{\mathrm{dlt}}$, including global brightness alternation, moving bright or dark patterns, and local intensity variations.

\paragraph{Training Data for Relighting Stage.}
In contrast, during the training of the relighting model, $V_{\mathrm{dlt}}$ is used as the input and $V_{\mathrm{rlt}}$ serves as the ground truth. 
For $\mathbf{D}_{\mathrm{Gen}}$, $V_{\mathrm{rlt}}$ is directly generated, while $V_{\mathrm{dlt}}$ is obtained by applying the trained delighting model to $V_{\mathrm{rlt}}$. 
Each relighting sample is trained with one active lighting condition selected from the available controls, while inactive branches are masked.
For $\mathbf{D}_{\mathrm{OLAT}}$, the available controls include $C_{\mathrm{e}}$ and $C_{\mathrm{v}}$; for $\mathbf{D}_{\mathrm{Gen}}^{\mathrm{rlt}}$, they include $C_{\mathrm{t}}$ and $C_{\mathrm{v}}$.

\section{Experiments}
\label{SecExper}

We evaluate \method\ as a unified framework for dynamic portrait video relighting. Our experiments are organized around the three lighting conditions: envmap-sequence-based relighting, lighting-text-based relighting, and lighting-reference-video-based relighting. 

\subsection{Experimental Setup}


\begin{figure*}[t]
    \centering
    \includegraphics[width=\linewidth]{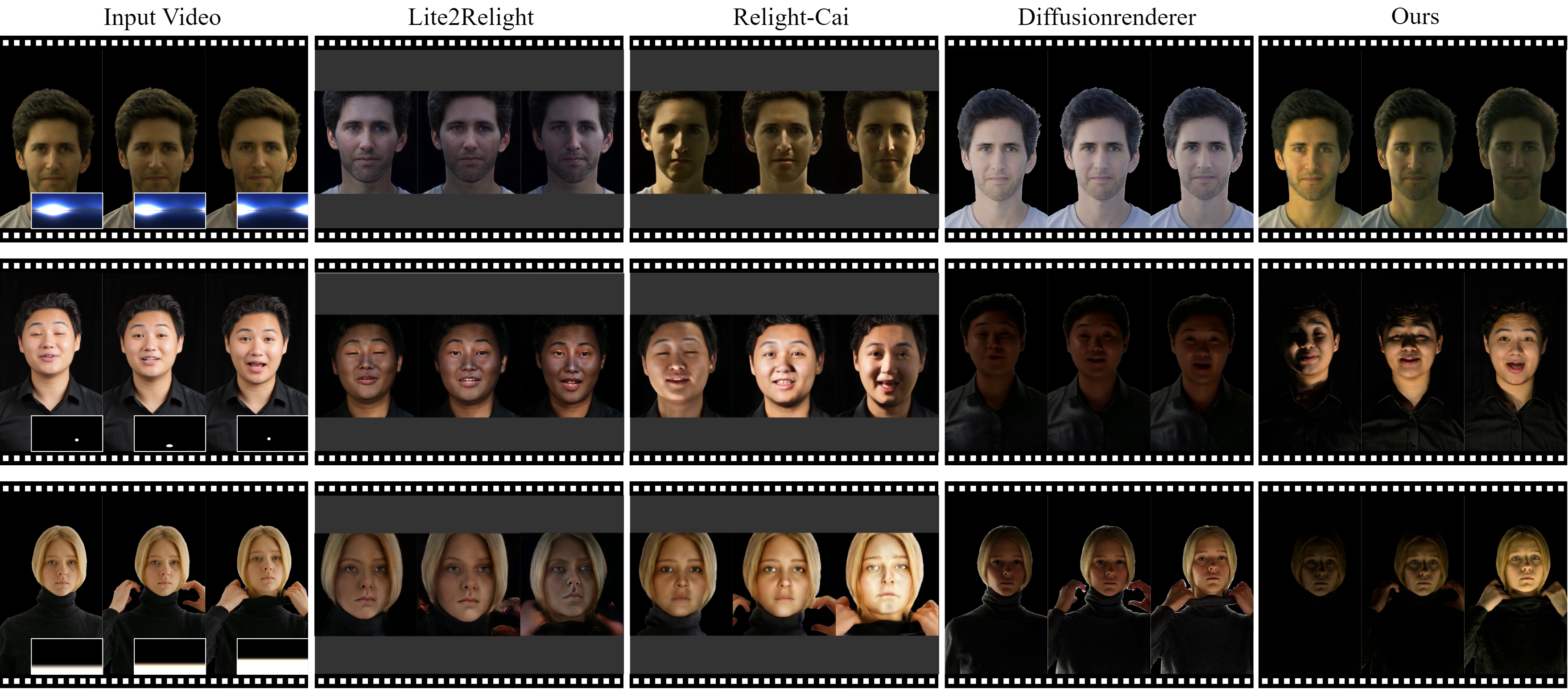}
    \caption{\textbf{Qualitative comparison for envmap-sequence-based relighting.}}
    \label{fig:envmap_qualitative}
\end{figure*}

\begin{table}[t]
\centering
\caption{Quantitative results for envmap-sequence-based relighting.}
\label{tab:envmap_quantitative}
\small
\setlength{\tabcolsep}{5pt}
\begin{tabular}{@{}lccccc@{}}
\toprule
\multirow{2}{*}{\textbf{Method}}
& \multicolumn{1}{c}{\textbf{Lighting Acc.}} 
& \multicolumn{3}{c}{\textbf{Visual Quality}} 
& \multicolumn{1}{c}{\textbf{Video Smoothness}} \\
\cmidrule(lr){2-2}
\cmidrule(lr){3-5}
\cmidrule(lr){6-6}
& \textbf{LE $\downarrow$} 
& \textbf{LPIPS $\downarrow$} 
& \textbf{ID $\uparrow$} 
& \textbf{PSNR $\uparrow$} 
& \textbf{tLPIPS $\downarrow$} \\
\midrule
Lite2Relight (SIGGRAPH'24)~\cite{Lite2Relight}
& 0.1423 & 0.4154 & 0.5653 & 15.6086 & 0.2721 \\
Relighting-Cai (CVPR'24)~\cite{relight-cai}
& 0.1585 & 0.4051 & 0.5193 & 14.2131 & 0.2684 \\
DiffusionRenderer (CVPR'25)~\cite{DiffusionRenderer}
& 0.1120 & 0.3781 & 0.6143 & 17.3683 & 0.2458 \\
\textbf{Ours} 
& \textbf{0.0584} & \textbf{0.1635} & \textbf{0.8097} & \textbf{23.1099} & \textbf{0.1303} \\
\bottomrule
\end{tabular}
\end{table}

\paragraph{Datasets.}
For quantitative evaluation of envmap-sequence-based relighting, we use the $\mathbf{D}_{\mathrm{OLAT}}$ test set, which contains 10 subjects and 16 envmap sequences that are not included in the training set.
For evaluating lighting-text-based relighting and lighting-reference-video-based relighting, we reserve 50 generated videos and collect 50 in-the-wild portrait videos from Pexels~\cite{Pexels}. None of these videos is used for training or fine-tuning \method.

\paragraph{Baselines.}
Our evaluation focuses on three tasks: envmap-sequence-based relighting, lighting-text-based relighting, and lighting-reference-video-based relighting. 
For envmap-sequence-based relighting, we use Lite2Relight~\cite{Lite2Relight}, Real-time 3D-aware Portrait Video Relighting~\cite{relight-cai}, denoted as Relighting-Cai, and DiffusionRenderer~\cite{DiffusionRenderer} as baselines. 
For lighting-text-based relighting, we compare with IC-Light~\cite{ICLight}, Light-A-Video~\cite{LightAVideo}, and Light-X~\cite{LightX}. Implementation details can be found in the Appendix \ref{app:baselines}.

\paragraph{Metrics.}
Following previous works~\cite{LuxPostFacto,LightX}, we evaluate paired envmap-sequence relighting using LE, LPIPS, ID similarity, tLPIPS, and PSNR. 
For lighting-text-based relighting, we use FID for overall visual quality, ID similarity for identity preservation, and tLPIPS for temporal stability.

\paragraph{Implementation Details.}
\method\ (including the delighting and relighting models) is built on Wan2.1-1.3B~\cite{wan2025}. 
We train the relighting model on 49-frame videos at a resolution of $832 \times 480$ for 40,000 steps using a three-stage training schedule, with a batch size of 8 on 8 H20 GPUs. 
Similarly, the corresponding delighting model is trained for 30,000 steps with a three-stage training schedule. 
More training details are provided in the Appendix \ref{app:training_setting}.

\subsection{Video Relighting Result}

\paragraph{Envmap-sequence-based relighting.}
Fig.~\ref{fig:envmap_qualitative} shows qualitative comparisons under envmap sequences. 
Frame-wise Lite2Relight and Relighting-Cai struggle to maintain temporal consistency, leading to flickering illumination and unstable appearance. 
DiffusionRenderer improves general relighting quality, but still fails to consistently preserve fine-grained temporal lighting evolution. 
In contrast, \method\ closely follows the target envmap sequence while preserving subject identity, motion, and temporally coherent shadows and highlights. 
Quantitative results in Table~\ref{tab:envmap_quantitative} further confirm these observations: \method\ achieves the best performance across all metrics.

\begin{figure*}[t]
    \centering
    \includegraphics[width=\linewidth]{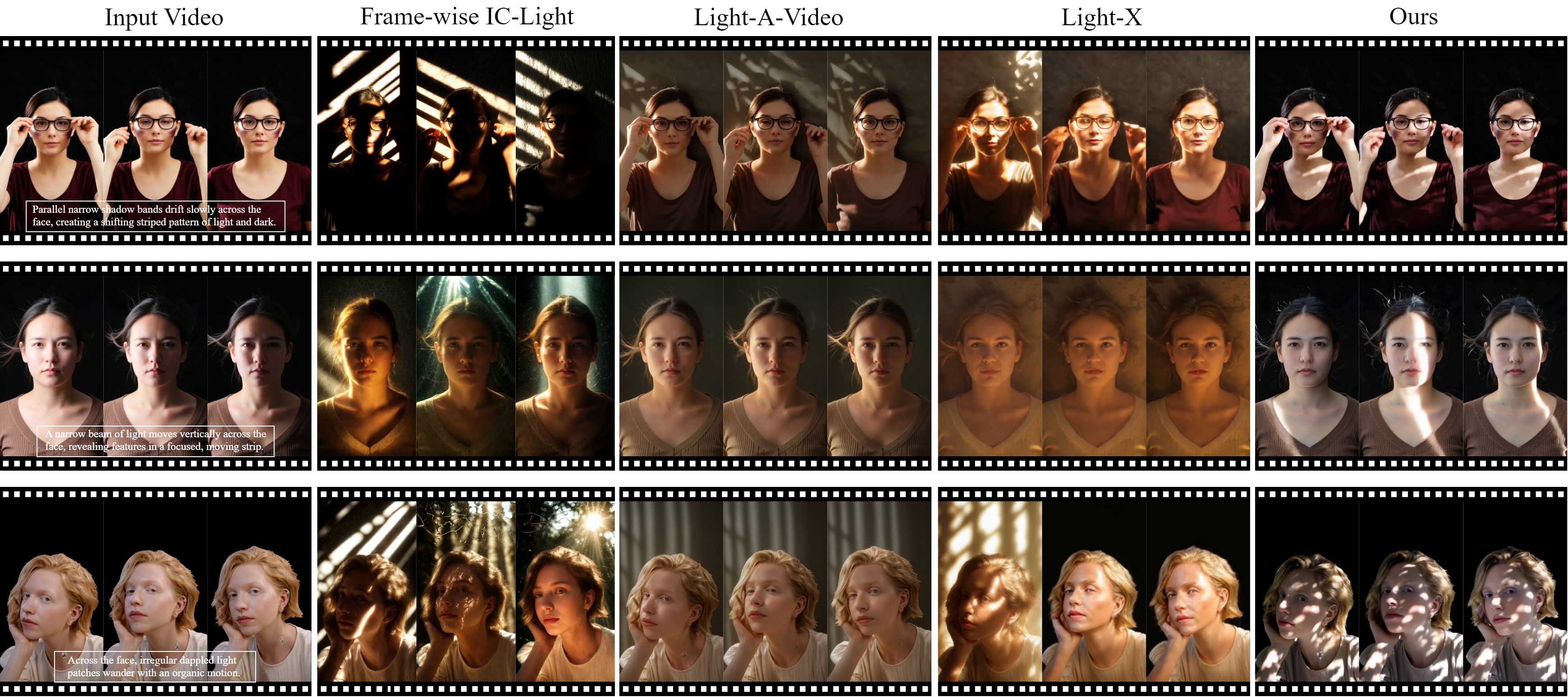}
    \caption{\textbf{Qualitative comparison for lighting-text-based relighting.}}
    \label{fig:text_qualitative}
\end{figure*}

\begin{wraptable}[9]{r}{0.44\columnwidth}
\vspace{-8pt}
\centering
\caption{Quantitative results for text-based relighting.}
\vspace{-3pt}
\label{tab:text_quantitative}
\small
\setlength{\tabcolsep}{3pt}
\begin{tabular}{@{}lccc@{}}
\toprule
\textbf{Method} 
& \textbf{FID $\downarrow$} 
& \textbf{ID $\uparrow$} 
& \textbf{tLPIPS $\downarrow$} \\
\midrule
IC-Light~\cite{ICLight}
& 134.98 & 0.4332 & 0.5715 \\
Light-A-Video~\cite{LightAVideo}
& 51.78 & 0.7126 & \textbf{0.1395} \\
Light-X~\cite{LightX}
& 66.27 & 0.6762 & 0.2224 \\
\textbf{Ours} 
& \textbf{31.16} & \textbf{0.7973} & 0.1417 \\
\bottomrule
\end{tabular}
\end{wraptable}
\paragraph{Lighting-text-based relighting.}
Fig.~\ref{fig:text_qualitative} shows qualitative comparisons for the evaluated dynamic lighting descriptions.
In these examples, frame-wise IC-Light exhibits flickering illumination and unstable appearance across frames.
Light-A-Video and Light-X leverage video priors, but show residual source-lighting effects and struggle to reproduce some of the target lighting changes.
\method\ uses the delighting stage to suppress source-lighting effects and then performs text-conditioned dynamic relighting with a DiT-based video model.
For the lighting descriptions evaluated here, the qualitative results suggest that \method\ better reproduces the illustrated lighting changes while preserving portrait identity and temporal coherence.
On this evaluation set, the quantitative results in Table~\ref{tab:text_quantitative} show that \method\ achieves the best FID and ID similarity, indicating better visual quality and identity preservation, while maintaining temporal stability comparable to the strongest video-based baseline.

\begin{figure*}[t]
    \centering
    \includegraphics[width=\linewidth]{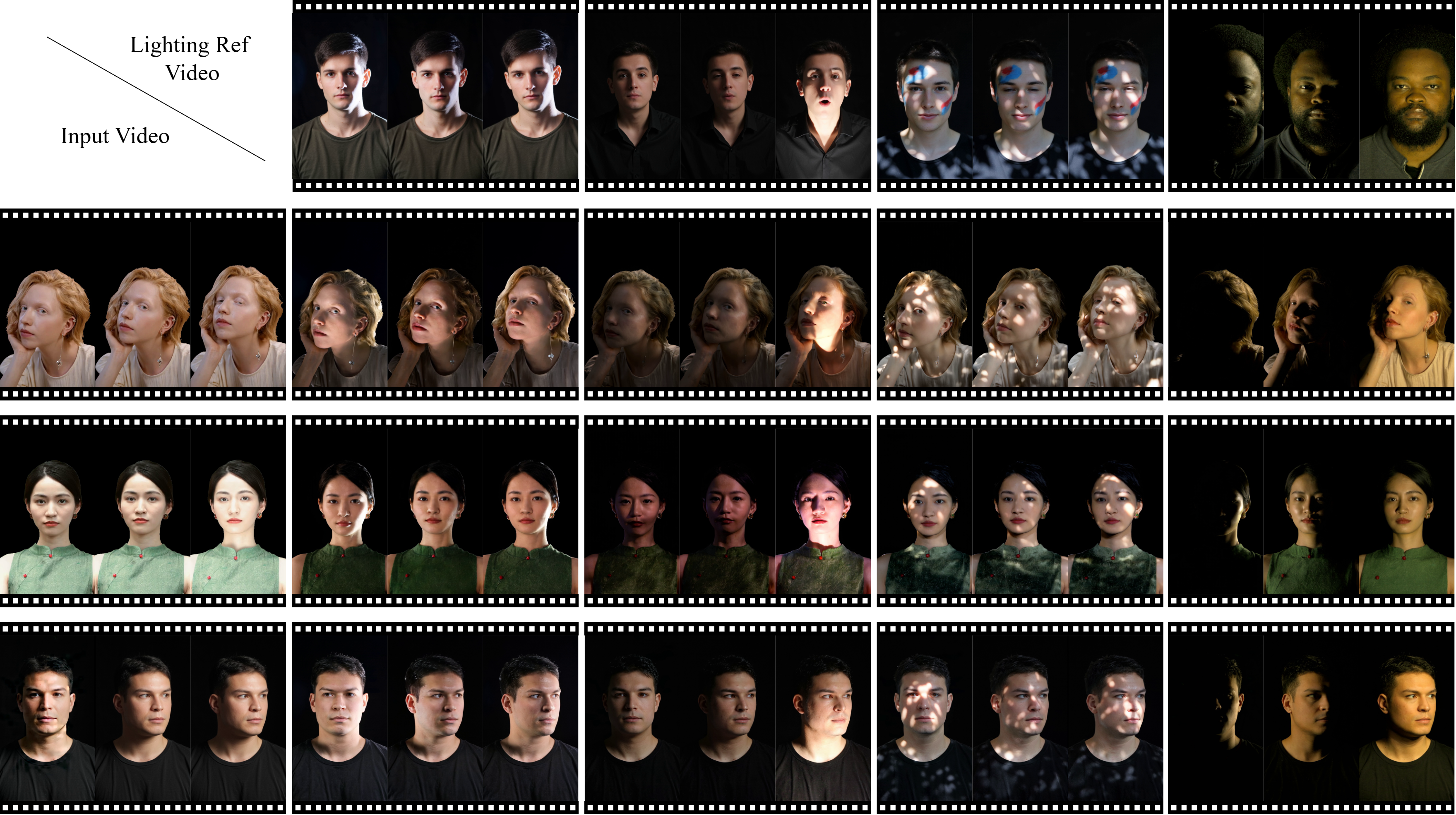}
    \caption{\textbf{Reference-video-guided lighting transfer.} \method\ transfers illumination dynamics from a lighting reference video while preserving the identity and motion of the source portrait video.}
    \label{fig:lighting_ref_result}
\end{figure*}

\paragraph{Lighting-reference-video-based relighting.}
As shown in Fig.~\ref{fig:lighting_ref_result}, \method\ supports reference-video-guided portrait relighting. 
Since paired ground truth is unavailable for this setting, we evaluate reference-video-guided lighting transfer qualitatively.
Given another portrait video as the lighting reference, \method\ transfers the dynamic illumination patterns from the reference video to the source video while preserving the source identity and motion.
Compared with text-based relighting, the lighting reference video provides more direct and fine-grained visual cues, including the temporal evolution of highlights, shadows, color changes, and flickering effects.
This enables more detailed dynamic lighting transfer across portrait videos.
Additional visual results for delighting, relighting, and cross-domain generalization are provided in the appendix.

\subsection{Ablation on Hybrid Training Data}

We conduct ablation studies to assess the contribution of our hybrid training data design in Sec.~\ref{sec:hybrid_training_data}. 
We compare three variants trained with different data sources: Static OLAT Data only $\mathbf{D}_{\mathrm{OLAT}}$, Dynamic Gen Data only $\mathbf{D}_{\mathrm{Gen}}$, and the full hybrid data $\mathbf{D}_{\mathrm{Hybrid}}$. 
The reported metrics are computed over the combined evaluation set covering three relighting settings: envmap-sequence-based relighting, lighting-text-based relighting, and lighting-reference-video-based relighting.

\begin{wrapfigure}[10]{r}{0.48\columnwidth}
\vspace{-40pt}
\centering
\includegraphics[width=\linewidth]{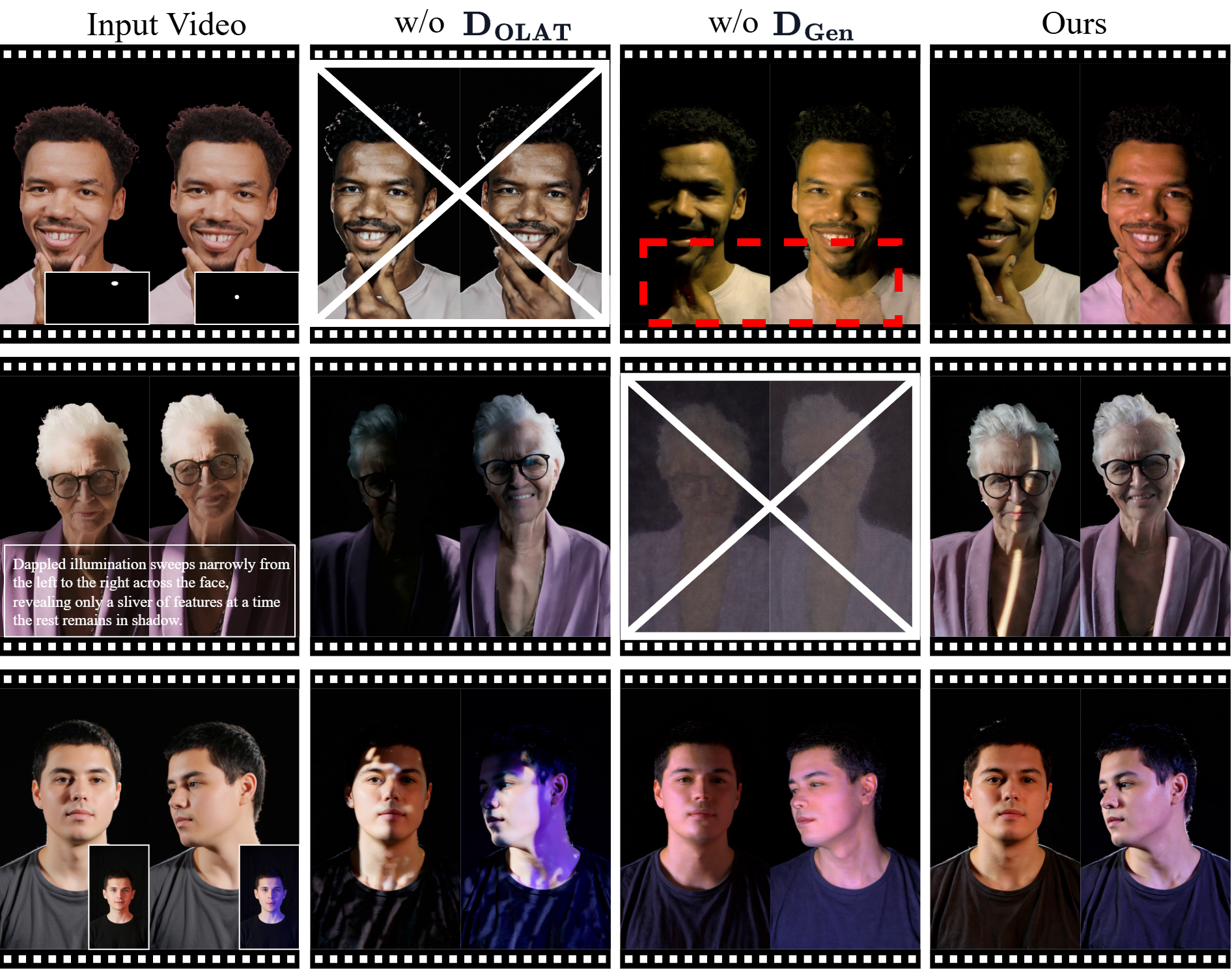}
\vspace{-12pt}
\caption{\textbf{Ablation on Hybrid Data.}}
\label{fig:ablation}
\vspace{-10pt}
\end{wrapfigure}

\noindent
\begin{minipage}{0.50\columnwidth}
\centering
\captionof{table}{Ablation study on hybrid training data.}
\label{tab:hybrid_data_ablation}
\vspace{-4pt}
\small
\setlength{\tabcolsep}{6pt}
\begin{tabular}{@{}ccccc@{}}
\toprule
$\mathbf{D}_{\mathrm{OLAT}}$ 
& $\mathbf{D}_{\mathrm{Gen}}$ 
& \textbf{FID $\downarrow$} 
& \textbf{ID $\uparrow$} 
& \textbf{tLPIPS $\downarrow$} \\
\midrule
& \cmark 
& 87.9401 & 0.5538 & 0.2380 \\
\cmark & 
& 86.9547 & 0.5470 & 0.2543 \\
\cmark & \cmark 
& \textbf{55.8078} & \textbf{0.6734} & \textbf{0.1906} \\
\bottomrule
\end{tabular}
\end{minipage}

\vspace{1pt}


As shown in Fig.~\ref{fig:ablation} and Table~\ref{tab:hybrid_data_ablation}, training with only $\mathbf{D}_{\mathrm{OLAT}}$ benefits from controlled relighting supervision, but struggles with complex portrait motion and diverse dynamic lighting effects.
Training with only $\mathbf{D}_{\mathrm{Gen}}$ improves motion and illumination diversity, but lacks the physically grounded constraints provided by OLAT data.
In contrast, the model trained on the hybrid data $\mathbf{D}_{\mathrm{Hybrid}}$ achieves the best overall performance across FID, ID similarity, and tLPIPS.
These results validate that controlled OLAT data and generated dynamic data provide complementary supervision for dynamic portrait video relighting.

\section{Conclusion}
\label{SecCon}

We introduce \method, a unified framework for portrait video relighting under dynamic lighting conditions.
The two-stage design first suppresses source illumination with a delighting model, and then leverages a DiT-based relighting model to synthesize relit videos under target lighting conditioned on environment-map sequences, dynamic lighting text, or lighting reference videos.
To enable training without costly dynamic OLAT capture or fully rendered paired datasets, we further construct a hybrid dataset that combines controlled static OLAT-based data $\mathbf{D}_{\mathrm{OLAT}}$ with scalable generated dynamic-lighting data $\mathbf{D}_{\mathrm{Gen}}$.
Extensive experiments demonstrate strong performance in environment-map-sequence relighting and reference-video-guided lighting transfer, while preserving portrait identity, motion, and temporal coherence.
Text-conditioned relighting is also demonstrated for the evaluated lighting descriptions.
This work provides a practical step toward flexible and controllable dynamic portrait relighting in real-world videos.

\section*{Acknowledgements}
We sincerely thank the anonymous area chair and reviewers for their insightful feedback.
We also thank vivo for providing computational resources.

This work was supported by research project KM20230861.

{
    \small
    \bibliography{main}
}

\newpage
\appendix
\section*{Appendix -- FlashRelight: Portrait Video Relighting with Dynamic Lighting}

\section{Supplementary Videos}
We provide additional video demos in the supplementary material.

\section{More Implementation Details}

\subsection{Training Setting of \method}
\label{app:training_setting}

Both the delighting model and the relighting model are initialized from Wan2.1-1.3B.
Unless otherwise specified, the remaining training hyperparameters follow the default setting of Wan2.1-1.3B.
During inference, we use 50 sampling steps with a shift value of 5.0.

\paragraph{Delighting model.}
The delighting model is trained with a three-stage schedule.
As shown in Table~\ref{tab:delighting_training_setting}, we first train the model on the static OLAT data $\mathbf{D}_{\mathrm{OLAT}}$ for 10,000 steps with a learning rate of $1\times10^{-4}$.
We then train it on the dynamic generated data $\mathbf{D}_{\mathrm{Gen}}$ for another 10,000 steps with a learning rate of $5\times10^{-5}$.
Finally, we fine-tune the model again on $\mathbf{D}_{\mathrm{OLAT}}$ for 10,000 steps with a learning rate of $1\times10^{-5}$.
The global batch size is 8.
We use LoRA fine-tuning with rank 16 and alpha 16, applying LoRA to the feed-forward network (FFN), self-attention, and cross-attention layers.

\begin{table}[h]
\centering
\caption{Training schedule of the delighting model.}
\label{tab:delighting_training_setting}
\small
\setlength{\tabcolsep}{10pt}
\begin{tabular}{@{}lccc@{}}
\toprule
\textbf{Stage} & \textbf{Training data} & \textbf{Steps} & \textbf{Learning rate} \\
\midrule
1 & $\mathbf{D}_{\mathrm{OLAT}}$ & 10,000 & $1\times10^{-4}$ \\
2 & $\mathbf{D}_{\mathrm{Gen}}$ & 10,000 & $5\times10^{-5}$ \\
3 & $\mathbf{D}_{\mathrm{OLAT}}$ & 10,000 & $1\times10^{-5}$ \\
\bottomrule
\end{tabular}
\end{table}

\paragraph{Relighting model.}
The relighting model is also trained with a three-stage schedule.
As shown in Table~\ref{tab:relighting_training_setting}, we first train the model on $\mathbf{D}_{\mathrm{OLAT}}$ for 10,000 steps with a learning rate of $1\times10^{-4}$.
We then train it on $\mathbf{D}_{\mathrm{Gen}}$ for 10,000 steps with a learning rate of $5\times10^{-5}$.
In the final stage, we fine-tune the model on the hybrid data $\mathbf{D}_{\mathrm{Hybrid}}$ for 20,000 steps with a learning rate of $1\times10^{-5}$.
The global batch size is 8.
We use LoRA fine-tuning with rank 32 and alpha 32, applying LoRA to the FFN, self-attention, and cross-attention layers.
In addition, the lighting projector and the newly added envmap cross-attention layers are trained with full-parameter optimization.

\begin{table}[h]
\centering
\caption{Training schedule of the relighting model.}
\label{tab:relighting_training_setting}
\small
\setlength{\tabcolsep}{10pt}
\begin{tabular}{@{}lccc@{}}
\toprule
\textbf{Stage} & \textbf{Training data} & \textbf{Steps} & \textbf{Learning rate} \\
\midrule
1 & $\mathbf{D}_{\mathrm{OLAT}}$ & 10,000 & $1\times10^{-4}$ \\
2 & $\mathbf{D}_{\mathrm{Gen}}$ & 10,000 & $5\times10^{-5}$ \\
3 & $\mathbf{D}_{\mathrm{Hybrid}}$ & 20,000 & $1\times10^{-5}$ \\
\bottomrule
\end{tabular}
\end{table}

\subsection{Inference Cost}
\label{app:cr_inference_cost}

Table~\ref{tab:cr_inference_cost} reports inference cost on a single NVIDIA H20 GPU for 49-frame clips at $832\times480$.
\method\ uses 50 sampling steps per stage, compared with 15 for the evaluated DiffusionRenderer configuration.
The \method\ timing covers generation calls for one measured clip after a warm-up clip, excluding external preprocessing, visualization, and video saving.
Memory is measured as PyTorch peak reserved GPU memory.

\begin{table}[t]
\centering
\caption{Stage-wise inference time and peak reserved memory. Total time sums both stages; peak memory is the maximum of their peaks.}
\label{tab:cr_inference_cost}
\small
\setlength{\tabcolsep}{5pt}
\begin{tabular}{@{}llcrr@{}}
\toprule
Method & Component & Steps & Time (s) & Memory (GiB) \\
\midrule
\method & Delighting & 50 & 87.5 & 15.17 \\
\method & Relighting & 50 & 98.4 & 15.23 \\
\method & Total & $50+50$ & 185.9 & 15.23 \\
DiffusionRenderer & Total & 15 & 503.3 & 44.22 \\
\bottomrule
\end{tabular}
\end{table}

Under these settings, \method\ takes 185.9\,s with a peak reserved memory of 15.23\,GiB.

\subsection{Implementation of Baselines}
\label{app:baselines}
All baselines are implemented using their official repositories and recommended inference settings.
For a fair comparison, the source videos are represented as 49-frame clips at $832\times480$ resolution before baseline-specific preprocessing.
For videos whose background is not pure black, we first segment the portrait foreground using BiRefNet-Portrait and use the resulting masks for foreground-aware evaluation.
When a baseline requires a different input resolution or aspect ratio, we apply the same geometric transformation to the source video, the ground-truth video, and the foreground mask, and map the baseline output back to the evaluation resolution when needed. All quantitative metrics are computed on the foreground regions.

\begin{itemize}
    \item \textbf{Lite2Relight.}
    Lite2Relight only supports $512\times512$ inputs.
    We therefore adapt each $832\times480$ frame to a $512\times512$ square input and run the method frame by frame.
    For quantitative evaluation, the ground-truth video and mask video are transformed with the same parameters as the source video, so all metrics are computed in the aligned $512\times512$ crop.

    \item \textbf{Relighting-Cai.}
    Relighting-Cai is also limited to $512\times512$ inputs.
    We use the same square adaptation protocol as Lite2Relight and apply the identical transformation to the ground-truth video and mask video for metric computation.
    Since this method takes low-order spherical harmonics as the lighting representation, we convert the target environment maps to SH coefficients before inference.
    The official implementation does not support colored relighting, and we therefore evaluate it under its supported monochromatic relighting setting.

    \item \textbf{DiffusionRenderer.}
    We use the Cosmos-Transfer1-DiffusionRenderer implementation.
    To match its required input size, we first pad the input frames with black borders and resize them to $1280\times704$.
    After relighting, we remove the padded region and resize the result back to $832\times480$ for visual comparison and quantitative evaluation.

    \item \textbf{IC-Light.}
    We use the official IC-Light implementation and apply it independently to each frame.
    The resulting frame-wise outputs are assembled back into 49-frame videos for evaluation.

    \item \textbf{Light-A-Video.}
    We use the official Light-A-Video implementation with its default video relighting pipeline.

    \item \textbf{Light-X.}
    Following the official implementation, we adapt each input video to $384\times672$ by black-border padding and resizing.
    After relighting, we crop the valid image region and resize the output back to $832\times480$.
\end{itemize}

\section{Details of Data Curation}

\subsection{Preprocessing of Static OLAT Data}
Static OLAT Data requires preprocessing of the OLAT observations, including alignment, color correction, and other steps. 
The detailed procedure can be found in the official FaceOLAT repository~\cite{3DPR}.

\subsection{VLM Prompt for Dynamic Gen Data Curation}
\label{VLM_prompt}

\begin{tcblisting}{promptbox,title={VLM Prompt for Lighting Extraction}}
    "You are a portrait-video dynamic lighting annotator. "
    "The input is a portrait video with dynamic lighting. "
    "Your task is to give one short but correct English description of the visible dynamic lighting on the portrait over time. "
    "Focus only on the lighting behavior on the face, hair, ears, jawline, chin, and neck. "
    "Use subject-relative positions such as front, front-left, front-right, left, right, top, and bottom. "
    "Assume the person faces the camera. "
    "\n\n"
    "The most important goal is to correctly describe: "
    "(1) where the visible light comes from relative to the person, "
    "(2) whether there are one or multiple light components, "
    "(3) the color of each light component, and "
    "(4) whether each light component is stable, flickering, pulsing, alternating, sweeping, or changing color over time. "
    "\n\n"
    "Always prioritize light position and flicker behavior. "
    "If a light is at the subject-front-right, say subject-front-right. "
    "If a light is at the subject-front-left, say subject-front-left. "
    "If one side has a stable warm light while another side has a colorful flickering light, describe them separately and clearly. "
    "Do not merge multiple distinct light components into one vague mixed description. "
    "If flicker is visible, mention it explicitly. "
    "If flashing is visible, mention it explicitly. "
    "If two sides flicker differently, describe both sides separately. "
    "If a light changes color over time, describe the color change in order. "
    "\n\n"
    "The description should be clear, and correct. "
    "Correctness is more important than detail. "
    "Output description of the dynamic lighting, with no JSON, no bullet points, and no extra explanation."
\end{tcblisting}

\newpage

\begin{tcblisting}{promptbox,title={Prompt for Lighting Text Enhancement}}
SYSTEM_PROMPT = (
    "You are a portrait-video dynamic-lighting text rewriter. "
    "Rewrite lighting descriptions into concise, diverse, natural English. "
    "Only describe the visible lighting behavior on the portrait. "
    "Preserve the lighting semantics exactly, including temporal change, direction, brightness, color, shadow pattern, and affected facial regions. "
    "Remove all dataset or capture constraints, such as invisible light source, off-camera source, black background, no objects, or frame/camera comments. "
    "Do not invent new lighting effects, objects, emotions, scenes, or camera motion. "
    "Do not output any thinking process, reasoning steps, analysis, explanation, or numbered list. "
    "Your response must contain exactly one line in this format: FINAL: <rewritten sentence>"
)

USER_PROMPT = (
    "/no_think\n\n"
    "Rewrite the following portrait lighting description into one natural English sentence.\n\n"
    "Core goal:\n"
    "- Make the wording more diverse and fluent.\n"
    "- Keep only the dynamic lighting content.\n"
    "- Preserve the original lighting meaning faithfully.\n\n"
    "Must preserve if present:\n"
    "- Temporal behavior: starts, gradually changes, flickers, pulses, sweeps, alternates, intensifies, fades.\n"
    "- Direction or position: front, side, left/right, above, below, overhead, under-face.\n"
    "- Brightness or contrast changes.\n"
    "- Color changes.\n"
    "- Shadow behavior and affected regions such as brows, eyes, nose, cheeks, chin, jawline, hair, or neck.\n\n"
    "Must remove:\n"
    "- Any mention that the light source is invisible, unseen, hidden, off-camera, or outside the frame.\n"
    "- Any mention of black background, empty background, no visible objects, no scene objects, or dataset constraints.\n"
    "- Any camera/frame wording unless it is necessary for the lighting itself.\n\n"
    "Diversity requirement:\n"
    "{style_instruction}\n\n"
    "Important diversity rules:\n"
    "- If the input is short or common, do not use the most obvious compressed rewrite.\n"
    "- Change the sentence structure, subject phrase, and main verb while preserving the lighting meaning.\n"
    "- Avoid returning a sentence that is nearly identical to the input.\n"
    "- Avoid generic rewrites such as: 'Irregular dappled light patches drift organically across the face.'\n"
    "- Avoid using the same wording as any forbidden previous rewrite listed below.\n\n"
    "Forbidden previous rewrites:\n"
    "{forbidden_outputs_block}\n\n"
    "Do not:\n"
    "- Do not add new light sources, colors, directions, objects, emotions, scene details, or camera motion.\n"
    "- Do not over-specify uncertain details.\n"
    "- Do not return JSON, bullets, quotes, explanations, reasoning, thinking process, or labels other than FINAL.\n\n"
    "Output format:\n"
    "FINAL: <one rewritten sentence>\n\n"
    "Now rewrite this description:\n"
    "{lighting_text}\n"
)
\end{tcblisting}

\begin{tcblisting}{promptbox,title={VLM Prompt for Quality Filtering}}
    "You are a strict portrait-video quality evaluator. "
    "Evaluate whether this video is suitable for portrait dynamic-lighting dataset training.\n\n"
    "Evaluation dimensions (score each from 0 to 10, one decimal place):\n"
    "1) background_clean_score: background is pure black / clean, no visible environment objects, and no light-beam spill on background.\n"
    "2) no_visible_light_source_score: no visible emitting source appears (no flashlight, flame, candle, bulb, tube, screen, neon sign, window light source, etc.).\n\n"
    "Output ONLY one JSON object with these keys exactly:\n"
    "background_clean_score, no_visible_light_source_score, "
    "has_visible_light_source, overall_score, pass, reasons.\n\n"
    "Rules:\n"
    "- background_clean_score, no_visible_light_source_score: decimal in [0,10], one decimal place.\n"
    "- Beam confinement constraint: light beams must stay on the person only. If a beam illuminates background, detaches from the person, or exists independently away from the person, treat as invalid for this dataset.\n"
    "- When beam confinement is violated, set background_clean_score<=3.0, overall_score<=60, pass=false, and explain in reasons.\n"
    "- has_visible_light_source: boolean.\n"
    "- overall_score: integer 0-100.\n"
    "- pass: boolean (true only when both dimension scores >= 8.0, has_visible_light_source=false, and overall_score >= 80).\n"
    "- reasons: concise English string, one sentence.\n"
    "- No markdown, no extra text outside JSON."
\end{tcblisting}

\section{More Experimental Results}

\subsection{Additional Results on Delighting}

\begin{figure*}[h]
    \centering
    \includegraphics[width=\linewidth]{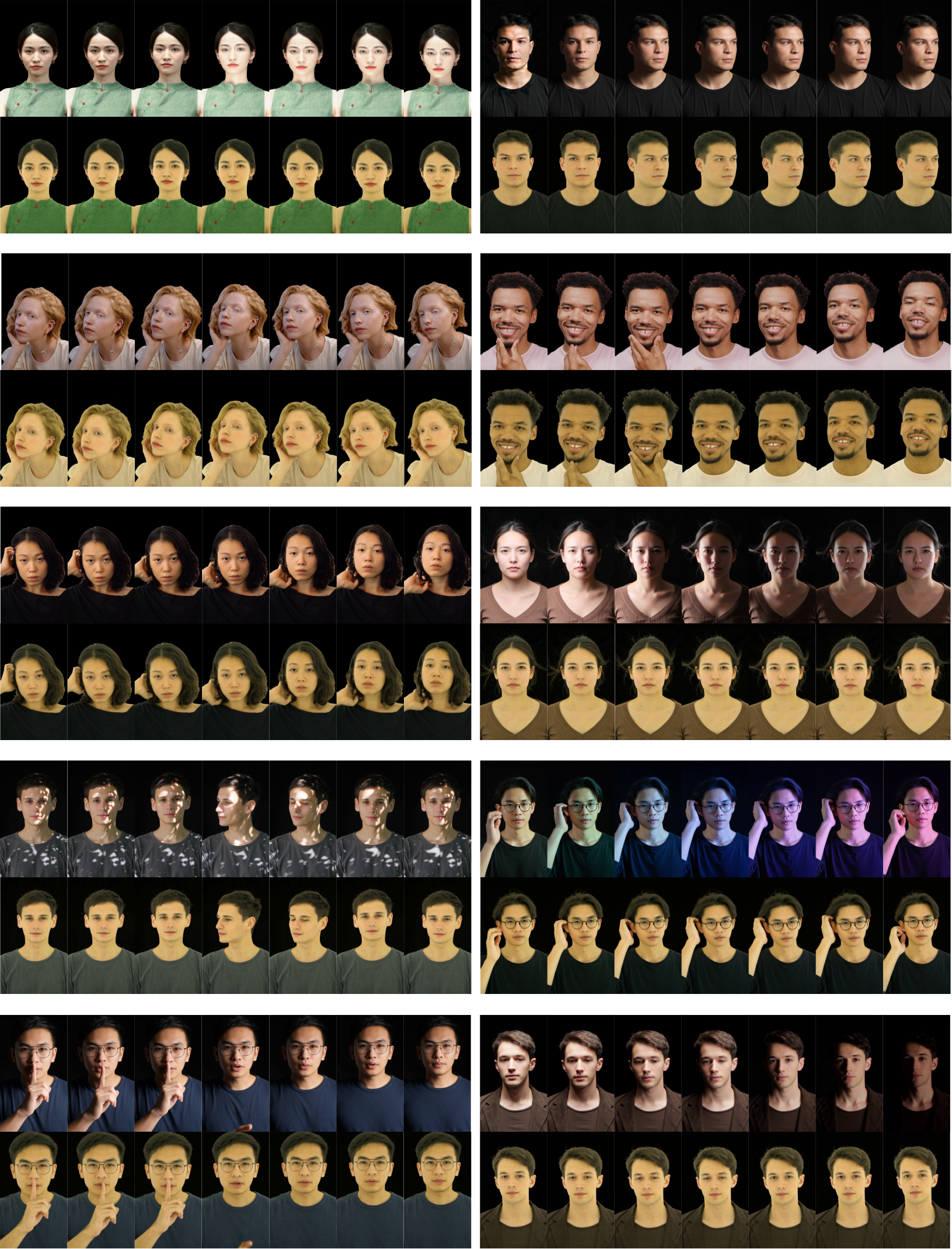}
    \caption{\textbf{Additional delighting results.} The first row shows source videos, and the second row shows the corresponding delighted results.}
    \label{fig:appendix_delight}
\end{figure*}

Fig.~\ref{fig:appendix_delight} shows additional results of the delighting stage.

\newpage
\subsection{Additional Results on Relighting}

\begin{figure*}[h]
    \centering
    \includegraphics[width=\linewidth]{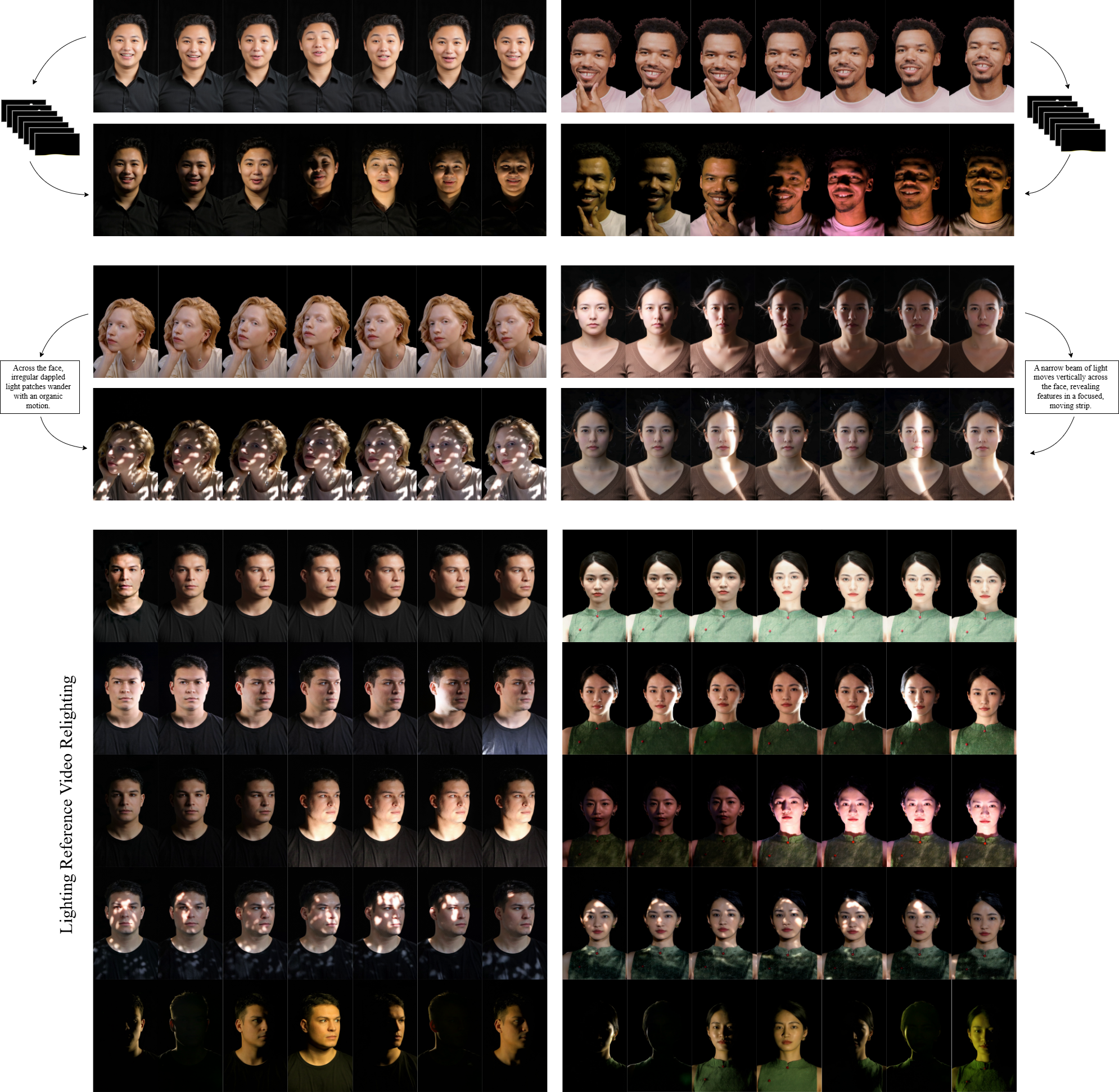}
    \caption{\textbf{Additional relighting results.} \method\ produces temporally coherent relit videos under diverse dynamic lighting conditions.}
    \label{fig:appendix_relight}
\end{figure*}

More relighting results are provided in the supplementary video.

\section{Generalization and Robustness}
\label{app:cr_control_generalization}

\subsection{Cross-Domain Generalization}
\label{app:cr_cross_domain}

We present visual results of \method\ on generated portrait videos%
\ifincludeadditionaldomains
, in-the-wild videos from Pexels, mobile-selfie videos from the \href{https://huggingface.co/datasets/UniqueData/selfie_and_video}{Selfies and Videos Dataset} (public Hugging Face samples), and digital-human clips extracted from \href{https://www.heygen.com/}{HeyGen demo videos}%
\else
\space and in-the-wild videos from Pexels%
\fi.
Figure~\ref{fig:cr_cross_domain} shows seven aligned input and relit frames per example, together with the corresponding target environment maps.

\begin{figure}[tbp]
\centering
\includegraphics[page=1,width=\linewidth]{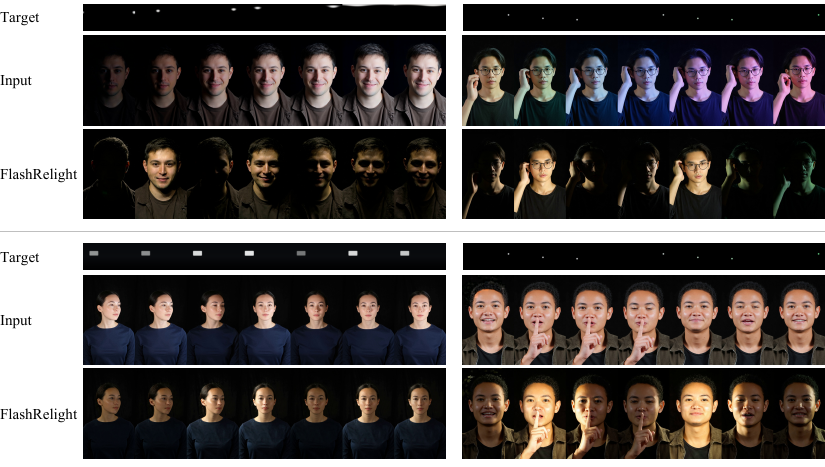}
\caption{\textbf{Cross-domain relighting: generated portrait videos.} Four examples are arranged in two columns and two rows. Each example shows seven aligned input and relit frames at $t=1,9,17,25,33,41,49$, together with the target environment maps. Target maps are tone-mapped with fixed exposure within each sequence.}
\label{fig:cr_cross_domain}
\end{figure}

\begin{figure}[tbp]
\ContinuedFloat
\centering
\includegraphics[page=2,width=\linewidth]{figures/pdf/cross_domain_examples.pdf}
\caption{\textbf{Cross-domain relighting (continued): in-the-wild portrait videos.} Four examples are arranged in two columns and two rows. Each example shows seven aligned input and relit frames at $t=1,9,17,25,33,41,49$, together with the target environment maps. Target maps are tone-mapped with fixed exposure within each sequence.}
\label{fig:cr_cross_domain_p02}
\end{figure}

\ifincludeadditionaldomains
\begin{figure}[tbp]
\ContinuedFloat
\centering
\includegraphics[page=3,width=\linewidth]{figures/pdf/cross_domain_examples.pdf}
\caption{\textbf{Cross-domain relighting (continued): mobile-selfie portrait videos.} Four examples are arranged in two columns and two rows. Each example shows seven aligned input and relit frames at $t=1,9,17,25,33,41,49$, together with the target environment maps. Target maps are tone-mapped with fixed exposure within each sequence.}
\label{fig:cr_cross_domain_p03}
\end{figure}

\begin{figure}[tbp]
\ContinuedFloat
\centering
\includegraphics[page=4,width=\linewidth]{figures/pdf/cross_domain_examples.pdf}
\caption{\textbf{Cross-domain relighting (continued): digital-human portrait videos.} Four examples are arranged in two columns and two rows. Each example shows seven aligned input and relit frames at $t=1,9,17,25,33,41,49$, together with the target environment maps. Target maps are tone-mapped with fixed exposure within each sequence.}
\label{fig:cr_cross_domain_p04}
\end{figure}

\fi

\subsection{Robustness to Dynamic and Extreme Lighting}
\label{app:cr_extreme_lighting}

\begin{figure}[tbp]
\centering
\includegraphics[page=1,width=\linewidth]{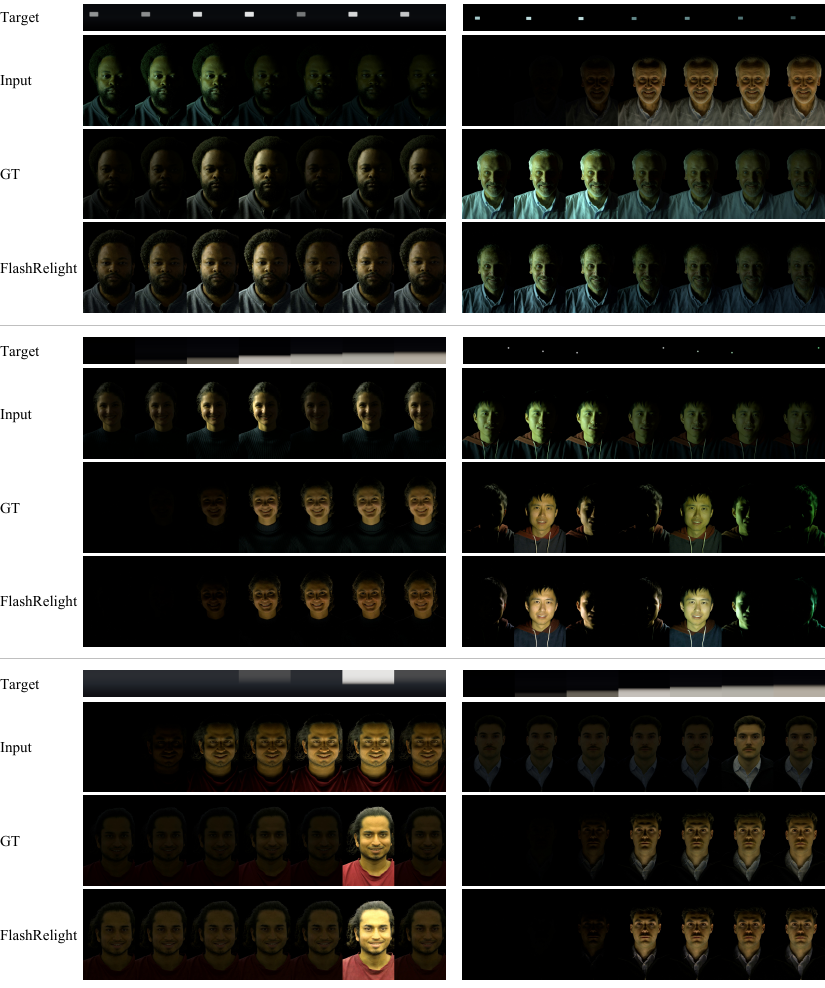}
\caption{\textbf{Relighting under dynamic illumination.} From left to right and top to bottom, the examples show flickering overhead light, TV-like lighting changes, uplighting, colored moving directional light, lightning-like illumination, and uplighting. Within each example, rows show target environment maps, input frames, paired ground truth (GT), and FlashRelight outputs. Seven aligned frames are shown at $t=1,9,17,25,33,41,49$, except the bottom-left lightning example, which uses $t=1,17,33,42,44,46,49$ to capture the flash. Target maps are tone-mapped with fixed exposure within each sequence.}
\label{fig:cr_extreme_examples}
\end{figure}

We present additional visual results under dynamic and extreme illumination, including moving directional light, overhead flicker, lightning-like flashes, TV-like lighting changes, and uplighting.
Figure~\ref{fig:cr_extreme_examples} shows seven aligned input, ground-truth, and relit frames per example, together with the target environment maps.
The paired ground-truth frames are constructed from OLAT captures.

\section{Limitations and Future Work}

\subsection{Limitations}
\method\ is limited by the accuracy of the intermediate delighting representation.
In particular, inputs with strong projected patterns, fast head motion, facial paint, motion blur, or severe image degradation can make source-light removal ambiguous.
In these cases, the delighting model may suppress not only illumination but also useful appearance cues such as skin tone, facial detail, or motion structure, and these errors can propagate to the relighting stage.
The hybrid data strategy also has finite coverage.
Although it avoids the cost of large-scale dynamic OLAT capture, the resulting training distribution is still shaped by the diversity of the available OLAT subjects, the design of the environment-map sequences, and the quality of the generated dynamic videos.
The current text-conditioned branch is effective for a limited set of lighting descriptions, but does not reliably follow general free-form lighting prompts.
The lighting captions produced by Qwen3.5-27B often do not accurately describe the illumination in the corresponding videos, introducing noise into text--video supervision.
These annotation inaccuracies may contribute to the limited generalization of text-based lighting control.
Finally, our current formulation primarily relights the portrait foreground.
Without explicit background relighting, foreground and background illumination may become inconsistent in unconstrained scenes.
The model can also lose temporal stability or fine shadow detail under severe occlusion and very rapid motion, and its diffusion-based implementation is better suited to offline high-quality editing than real-time use.

\subsection{Future Work}
One natural extension is to build \method\ on stronger video foundation models, such as newer Wan-series models, which may provide better motion priors and more faithful long-range lighting dynamics.
Another important direction is scene-level relighting, where the foreground portrait and background are edited jointly to maintain global illumination consistency.
On the data side, richer lighting programs and more diverse portrait captures could improve data coverage.
A promising direction is to use more capable video-language
models to improve the annotation of lighting attributes and
their temporal evolution.
Rebuilding the text--video training pairs with more accurate
annotations and more diverse lighting descriptions could
improve text-conditioned relighting and its generalization
to free-form lighting prompts.
The robustness of the system could also be improved by explicitly training on blur, occlusion, accessories, extreme expressions, and non-standard facial appearances.
Finally, model distillation, temporal feature caching, and lightweight video diffusion backbones are promising routes toward interactive or real-time portrait relighting while retaining the visual quality of the current offline system.

\section{Societal Considerations}

\method\ targets authorized portrait video editing, where controllable lighting can make post-production and virtual production workflows more accessible.
Since the method can alter how a person and scene are perceived, inappropriate use may obscure the original capture conditions, modify a subject's appearance without consent, or make edited footage harder to distinguish from an unmodified recording.
We therefore encourage use on consented or properly licensed videos, clear disclosure when relighting has been applied, and provenance support such as metadata, watermarking, or forensic detection tools for public-facing content.

\clearpage

\end{document}